\documentclass[final]{cvpr}

\usepackage{times}
\usepackage{epsfig}
\usepackage{graphicx}
\usepackage{amsmath}
\usepackage{amssymb}

\usepackage{array}
\usepackage{tabulary}
\usepackage{multirow}
\usepackage[table]{xcolor}
\usepackage{ctable}
\usepackage{booktabs}		
						
\usepackage{kotex}
\usepackage[pagebackref=true,breaklinks=true,colorlinks,bookmarks=false]{hyperref}

\def\cvprPaperID{****} 
\def\confYear{CVPR 2021}

\begin{document}

\title{Mounting Video Metadata on Transformer-based Language Model for Open-ended Video Question Answering}

\author{Donggeon Lee\\
Seoul National University\\
{\tt\small dglee@bi.snu.ac.kr}

\and
Seongho Choi\\ 
Seoul National University\\

{\tt\small shchoi@bi.snu.ac.kr}

\and
Youwon Jang\\
Seoul National University\\

{\tt\small ywjang@bi.snu.ac.kr}

\and
Byoung-Tak Zhang\\
Seoul National University\\

{\tt\small btzhang@bi.snu.ac.kr}
}

\maketitle

\begin{abstract}
Video question answering has recently received a lot of attention from multimodal video researchers. Most video question answering datasets are usually in the form of multiple-choice. But, the model for the multiple-choice task does not infer the answer. Rather it compares the answer candidates for picking the correct answer. Furthermore, it makes it difficult to extend to other tasks. In this paper, we challenge the existing multiple-choice video question answering by changing it to open-ended video question answering. To tackle open-ended question answering, we use the pretrained GPT2 model. The model is fine-tuned with video inputs and subtitles. An ablation study is performed by changing the existing DramaQA dataset to an open-ended question answering, and it shows that performance can be improved using video metadata.
\end{abstract}

\section{Introduction}
Transformers are now the de facto standard for language modeling and recently extending their applications in vision and multimodal domain~\cite{vaswani2017, chen2020uniter}. Transformers in the vision and language domain are usually pretrained with large-scale datasets and applied to various downstream tasks.
Among downstream tasks, video question answering evaluates whether the model understands various dimensions of video contents and is usually done in multiple-choice.
However, when learning a model for multiple-choice video question answering, the model selects the correct answer by comparing the similarity between the question and the answer candidates rather than inferring the correct answer to the question.
But, selecting the correct answer through comparison with the answer candidates does not perform the reasoning required in the question and answering, making it difficult to generalize for other tasks.

In this paper, we tackle the current multiple-choice video question answering dataset by changing it into an open-ended format. The answer candidates are not given in open-ended multimodal video question answering, so the model infers the correct answer through reasoning. In addition, it is possible to develop a model that can be applied to other tasks except for the decoder part that generates the correct answer.

Challenging open-ended multimodal video question answering, we propose an extended model that learns various modalities together based on the recently proposed Transformer language model. The proposed model receives various metadata and language input of video.
The results show that performance can be improved by combining multiple metadata rather than features from raw videos.

This paper is organized as follows. Chapter 2 examines related works to video question answering and open-ended question answering. Chapter 3 describes the proposed model and learning strategy. Chapter 4 examines the dataset and experimental settings, as well as the quantitative results. Finally, in Chapter 5, the conclusion and future research directions are described.

\section{Related Work}
\subsection{Video Question Answering}
A variety of video question-answering datasets have been proposed, including MovieQA\cite{tapaswi2016movieqa}, PororoQA\cite{kim2017deepstory}, TGIF-QA\cite{jang2017tgif}, TVQA\cite{lei2019tvqa}, DramaQA\cite{choi2020dramaqa}, and are mostly in the multiple-choice format. AVSD Dataset\cite{alamri2019audiovisual} is characterized by the fact that question-answering for video is in the form of dialogue, which is out of the existing multiple-choice form.

Recently, various approaches have been proposed for video story question answering, which can be divided into three categories. There are techniques using Memory Network\cite{tapaswi2016movieqa, kim2017deepstory}, Attention\cite{kim2017deepstory, lei2019tvqa}, and Transformer\cite{yang2020bert}. Memory networks stores and utilizes key information about a question-answering in a memory network to find it among many information in a long video.  
Attention effectively represents only the representation of visual/verbal core information by progressing attention across layers.
Techniques utilizing context matching by applying attention achieved high performance in question-and-answer by comparing the context of a question-and-answer with the context of a given video in detail.
Recently, researchers propose transformer-based models for video question answering. \cite{vaswani2017attention} proposed transformer and the proposed architecture brought a huge performance improvement in language modeling, and there is a move to expand it to a video domain. Recent state-of-art models show that these techniques can perform well in modeling the video as well as the language.

\subsection{Opend-Ended Question Answering}

In the H. Xue et al.\cite{xue2017unifying}, Z. Zhao et al.\cite{zhao2018open}, pointed out that the existing video question answering task used only one static image and text and also dealt with it as a short-word-oriented multiple-choice problem. It is emphasized that this approach cannot utilize the sequential and temporal information of the video. Therefore, its usability is limited in that the answer is chosen within given answers. In the above papers, the sequential/time information of the video was utilized to finally generate answers through decoders, resulting in better results than traditional methods (Mean-VQA, SS-VQA, etc.). However, the issues addressed by the above papers are limited in that they are short-lived, although open-ended, and the format of questions and answers is also simple.

In the \cite{li2020bridging}, the author conducted a study on AVSD task\cite{alamri2019audiovisual}(Given video and ten turns of question answering a text, task generates natural language answers to the last question) based on Transformer(GPT2\cite{radford2019language}). This paper extracts features from video and text with I3D\cite{carreira2017quo} and VGGish\cite{hershey2017}, applies positional encoding, Beam Search, receives good results from several metrics (BLEU, METEOR, CIDEr, etc.). However, the model is not much different from B, and the position and video feature information was not used properly.




\begin{figure*}[!h]
\centering
\includegraphics[scale=0.5]{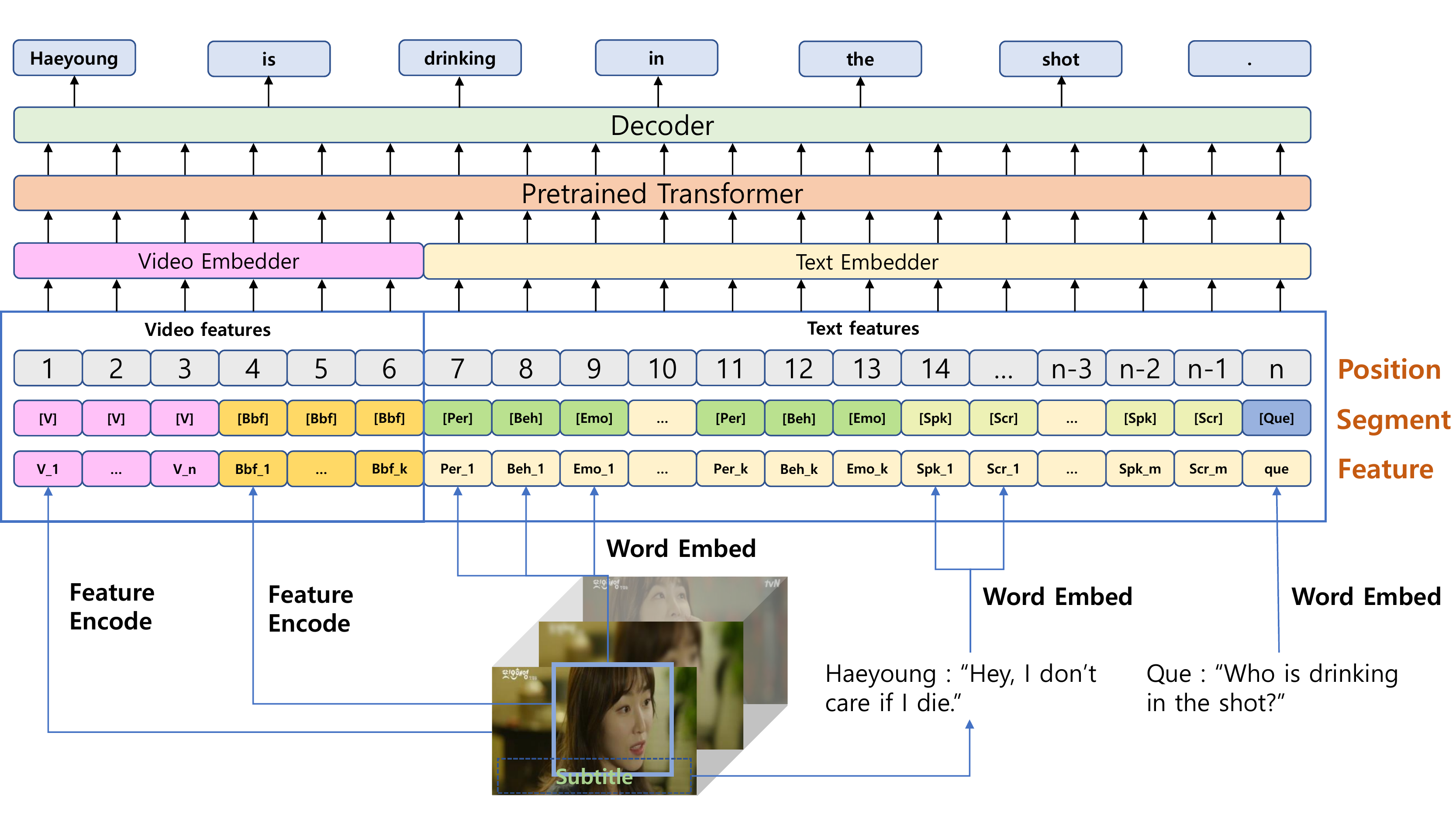}
\caption{Multimodal transformer model architecture. The video embedder is a linear layer which embeds feature of video size to feature of embedding size, and the text embedder is a linear layer which embeds feature of vocab size to feature of embedding size.  denote We used the following segment tokens [V] : Video, [Bbf] : feature of bounding box, [Per] : person's name, [Beh] : person's behavior, [Emo] : person's emotion, [Spk] : speaker, [Scr] : script, [Que] : question. }
\label{fig:model}
\end{figure*}

\section{Method}
 
\subsection{Formulation}
 The purpose of our model is to integrate multimodal information (e.g., subtitle, video, audio, question, etc.) to generate the open-ended answer.
 
 Our model consists of inputs of video, question and outputs of answer. The video is represented as $\mathbf{V}=(\{\mathbf{v}_{1}, \ldots , \mathbf{v}_N\}, \{\mathbf{m}_1, \ldots , \mathbf{m}_N\},  \{\mathbf{s}_1, \ldots , \mathbf{s}_M\})$. $\mathbf{v}_n$is representing the n-th frame in $\mathbf{V}$, $\mathbf{m}_n$ means a image features, and a visual meta data, the information such as person, person's emotion and behavior, in bounding box corresponding to n-th frame, $\mathbf{s}_m$ is m-th subtitle in the entire video $\mathbf{V}$. The question is represented as $\mathbf{Q}=\{w_a^1, \ldots, w_q^L\}$, and the answer is represented as $\mathbf{A}=\{w_a^1, \ldots, w_a^K\}$.
 
 Each frame can be expressed as $v_{\mathbf{v}_n}$ by extracting 3 frames per second from video and then feeding in the pre-trained I3D\cite{carreira2017quo} model to extract feature vectors.

 There is information about the character in the form of $\{\mathbf{c}^1_{\mathbf{v}_n}, \ldots, \mathbf{c}^{I_{\mathbf{m}_n}}_{\mathbf{m}_n}\}$
 in each $\mathbf{m}_n$. and information about each character is represented as $\mathbf{c}^i_{\mathbf{m}_n} = (f^i_{\mathbf{m}_n}, p^i_{\mathbf{m}_n}, b^i_{\mathbf{m}_n}, e^i_{\mathbf{m}_n})$.
 
 $f^i_{\mathbf{m}_n}$ is a feature representation of the character's image of bounding box using a pre-trained ResNet152\cite{he2016deep} model. $p^i_{\mathbf{m}_n}$ is a word embedding representation using a pre-trainned GPT2 model. $b^i_{\mathbf{m}_n}$ is the character's behavior. $e^i_{\mathbf{m}_n}$ is a word embedding representation of the character's emotion.
 
 Each $\mathbf{s}_m$ an be expressed as $(p_{\mathbf{s}_m}, \{w_{\mathbf{s}_m}^1,\ldots,{w}_{\mathbf{s}_m}^{J_{\mathbf{s}_m}}\})$ which which can be divided into sentence, $\{{w}_{\mathbf{s}_m}^1,\ldots,{w}_{\mathbf{s}_m}^{J_{\mathbf{s}_m}}\}$, which can be divided into a word ${w}_{\mathbf{s}_m}^j$ and a speaker ${p}_{\mathbf{s}_m}$
  Both speakers and words can be expressed in a previous way. Sentences can also be broken down into words using the GPT2 tokenizer.

\subsection{GPT2}
  We reference and use GPT2, a transformer model, which uses attention in place of the previous recurrence- and convolution-based architectures. Attention mechanisms allow the model to selectively focus on segments of input text it predicts to be the most relevant.

  GPT2 models receive the feature, segment, and position as inputs. Feature refers to data that embeds text input through GPT2 tokenizer, segment refers to data that means a token type of each word, such as <eos> and <sos>, and position refers to the location of each word in the sentence.

\subsubsection{Feature Embedding}
 
 Feature embedding input is all of the preceding $(v_{\mathbf{v}_n}, \{\mathbf{c}^1_{\mathbf{m}_n}, \ldots, \mathbf{c}^{I_{\mathbf{m}_n}}_{\mathbf{m}_n}\})$ to a two-dimensional sequence over time. Subsequent $({p}_{\mathbf{s}_m}, \{{w}_{\mathbf{s}_m}^1,\ldots,{w}_{\mathbf{s}_m}^{J_{\mathbf{s}_m}}\})$ similarly leads to a two-dimensional sequence over time. Finally, we attach $\{w_q^1, \ldots, w_q^L\}$. Therefore, the sequence length is $N + \sum_{{\mathbf{m}_n}=1}^N I_{\mathbf{m}_n} + M+ \sum_{{\mathbf{s}_m}=1}^M J_{\mathbf{s}_m} +L$. On the other hand, if features are extracted using I3D or ResNet, the features are different from those extracted with GPT2 models, so the dimensions are adjusted through a layer of learnable linear layers.

\begin{align*}
    \mathbf{V}_{feature}= [&\{(v_{\mathbf{v}_n}, \{\mathbf{c}^1_{\mathbf{m}_n}, \ldots, \mathbf{c}^{I_{\mathbf{m}_n}}_{\mathbf{m}_n}\})\}, 
    \\& \{({p}_{\mathbf{s}_m}, \{{w}_{\mathbf{s}_m}^1,\ldots,{w}_{\mathbf{s}_m}^{J_{\mathbf{s}_m}}\})\}, 
    \\& \{qw_1, \ldots, qw_L\}]
\end{align*}

\subsubsection{Segment Embedding}

\begin{table}[h!]

\begin{tabular}{p{0.15\columnwidth} | p{0.7\columnwidth}} 

\specialrule{.1em}{.05em}{.05em} 
    Notation & Description\\
\specialrule{.1em}{.05em}{.05em}     
    [V] & I3D feature for each frame \\
    \hline
    [BBF] & 2D ResNet feature for each bounding box \\
    \hline
    [PER] & Name of each character \\
    \hline
    [BEH] & Behavior of each character\\
    \hline
    [EMO] & Emotion of each character \\
    \hline
    [SPK] & Speaker of each subtitle\\
    \hline
    [SCR] & Each subtitle \\
    \hline
    [QUE] & Question \\
\specialrule{.1em}{.05em}{.05em} 
\end{tabular}
\centering
\caption{Notation and description for segments.}
\label{table:segment}
\end{table}

Segment embedding distinguishes the various inputs that enter the video. The distinguishing features can be divided into eight as Table \ref{table:segment}.

For each of these eight Feature categories, Segment embedding was performed using special token in GPT2.
 
\begin{table}[h!]

\begin{tabular}{l @{\hskip 0.1in} c @{\hskip 0.1in} c @{\hskip 0.1in} c @{\hskip 0.1in} c @{\hskip 0.1in} c @{\hskip 0.1in} @{\hskip 0.1in} c} 

\specialrule{.1em}{.05em}{.05em} 
    Method
     & Bleu
     & Meteor
     & Bertscore
     & Bleurt
     & Time\\
     
\specialrule{.1em}{.05em}{.05em}     
    
    \hline
    Beam &0.69	&0.2	&0.34	&0.62 &8 min\\
    \hline
    Nucleus &0.68	&0.18	&0.32	&0.6 &130 min\\
    
\specialrule{.1em}{.05em}{.05em} 
\end{tabular}
\centering
\caption{It is a description of the performance and time required for each Decoding Method for 4385 data in a subtitle-only environment.}
\label{table:quan_result}
\vspace{-1.0em}
\end{table}

\subsection{Decoding Method}

To find an effective decoding method for multimodal answer generation, we try the decoding methods, including beam search and Nucleus Sampling\cite{holtzman2019curious} which samples text from the dynamic nucleus of the probability distribution. Although beam search showed slightly high performance, it took about 16 times more time to use it in real-time, so Neclues Sampling was used.

\subsection{Implementation Details}

All experiments are run on NVIDIA [TITAN Xp]. Because of the lack of memory, we use a batch size of 1 input unit. We use AdamW optimizer\cite{loshchilov2017decoupled} with a learning rate of 1e-4 and weight decay of 1e-5. Cross-entropy loss is used to train the model.

\section{Results}

\subsection{Evaluation}

The evaluation is carried out using BLEU\cite{papineni2002bleu} based on n-gram, METEOR\cite{banerjee2005meteor} considering recall as a traditional metric to evaluate the generated text. In addition, we evaluate the answers generated with a total of four metrics, including BERTScore\cite{zhang2019bertscore} which is measured based on a similarity between each token embedding and BLEURT\cite{sellam2020bleurt} which uses the pre-learned model as metric.

\subsection{Quantitative Results}

\begin{table}[ht!]

\begin{tabular}{l @{\hskip 0.1in} c @{\hskip 0.1in} c @{\hskip 0.1in}c @{\hskip 0.1in} c @{\hskip 0.1in} @{\hskip 0.1in} c} 
\specialrule{.1em}{.05em}{.05em} 
    Model
     & Bleu
     & Meteor
     & Bertscore
     & Bleurt \\
     
\specialrule{.1em}{.05em}{.05em}     
    
    \hline
    
    S & 0.68	&0.18	&0.32	&0.6\\
    \hline
    S + V &0.65	&0.1	&0.3	&0.59 \\
    S + B &0.697	&0.202	&0.35	&0.6\\
    
    \hline
    S + M & 0.733	&0.281&	0.378	&0.62\\
    S + M, V & 0.726	&0.263	&0.38	&0.61 \\
    S + M, B & 0.733	&0.276	&0.38	&0.62 \\
    S + M, V, B & 0.724	&0.258	&0.37	&0.61 \\
    
\specialrule{.1em}{.05em}{.05em} 
\end{tabular}
\centering
\caption{Quantitative experimental results for the DramaQA validation set. \textbf{S} stands for subtitle, \textbf{V} stands for video features extracted from I3D, \textbf{B} stands for bounding box features extracted from ResNet, and \textbf{M} stands for visual metadata composed of person, emotion, and behavior.}
\label{table:quan_result}
\vspace{-1.0em}
\end{table}
Table \ref{table:quan_result} shows metadata plays a major role in improving performance. Our model is based on GPT2, so there is language bias. It helps improve performance with language metadata.

The information in bounding box features also helps answer questions by looking at S / B + S. However, comparing M + S / B, M + S did not improve performance.

Video information lowers performance. For reasons, a transformer-based model is a model with large language bias, and the entire video that is irrelevant to the question works even worse than bounding box features.
    
\section{Conclusion}
In this paper, we challenge the existing multiple-choice video question answer by converting it into an open-ended form. We construct the model in the form of a multimodal transformer by adding video and metadata from video to the existing pre-trained language model. Ablation studies using the DramaQA dataset showed that video metadata helped performance.

For future work, we plan to use the dense caption features in the video space transferred into the language space to circumvent the language bias problem. 

{\small
\bibliographystyle{ieee_fullname}
\bibliography{egbib}
}

\end{document}